\pdfoutput=1

\documentclass{article}
\usepackage{ijcai21}

\usepackage{times}
\usepackage{soul}
\usepackage{url}
\usepackage[hidelinks]{hyperref}
\usepackage[utf8]{inputenc}
\usepackage[small]{caption}
\usepackage{graphicx}
\usepackage{amsmath}
\usepackage{amsthm}
\usepackage{booktabs}
\usepackage{algorithm}
\usepackage{algorithmic}
\usepackage{color}

\usepackage{amsfonts}

\def\eg{e.g.}
\def\ie{i.e.}
\def\etc{\textit{etc.}}

\title{PEARL: Parallelized Expert-Assisted Reinforcement Learning \\ for Scene Rearrangement Planning}

\author{
Hanqing Wang$^1$\and
Zan Wang$^1$\and
Wei Liang$^1$\And
Lap-Fai Yu$^2$
\affiliations
$^1$Beijing Institute of Technology\\
$^2$George Mason University
\emails
\{hanqingwang, wangzan, liangwei\}@bit.edu.cn,
craigyu@gmu.edu
}

\begin{document}

\maketitle

\begin{abstract}
  \textit{Scene Rearrangement Planning (SRP) is an interior task
    proposed recently. The previous work defines the action space of
    this task with handcrafted coarse-grained actions that are
    inflexible to be used for transforming scene arrangement and
    intractable to be deployed in practice. Additionally, this new
    task lacks realistic indoor scene rearrangement data to feed
    popular data-hungry learning approaches and meet the needs of
    quantitative evaluation. To address these problems, we propose a
    fine-grained action definition for SRP and introduce a large-scale
    scene rearrangement dataset. We also propose a novel learning
    paradigm to efficiently train an agent through self-playing,
    without any prior knowledge. The agent trained via our paradigm
    achieves superior performance on the introduced dataset compared
    to the baseline agents. We provide a detailed analysis of
    the design of our approach in our experiments.}
\end{abstract}

\section{Introduction}
Rearrangement is a task of bringing a physical environment into a
specific goal state, leading to practical applications such as setting
tables, packing objects, and rearranging furniture. To automate this
task, an agent is called upon to perform actions such as analysis,
planning, and moving objects. There has been a rich body of studies on
rearrangement planning in robotics, for example, devising a robotic
arm to rearrange objects on a tabletop through grasping, picking, or
nonprehensile
actions~\cite{labbe2020monte,yuan2019end,koval2015robust}.

Recently, an emerging task of automatic \textit{Scene Rearrangement
  Planning} (SRP), filling the gap between scene synthesis and layout
realization, draws attention from
researchers~\cite{wang2020scenem,Xiong2020,Batra}. In SRP, given an
initial scene layout and a target scene layout, the goal is to find a
feasible move plan by which an agent can move furniture and transform
an initial layout to a target layout. Compared with the conventional
tabletop rearrangement task, SRP is more complicated because SRP needs
to consider the factors of object poses, object shapes, and terrain
constraints, which are usually simplified in a tabletop setting to
make the problem tractable. On the other hand, the decision sequence
of SRP is often longer due to the large scale and complexity of
scenes, making the move planning task more challenging.

\begin{figure}[t]
  \includegraphics[width=\linewidth]{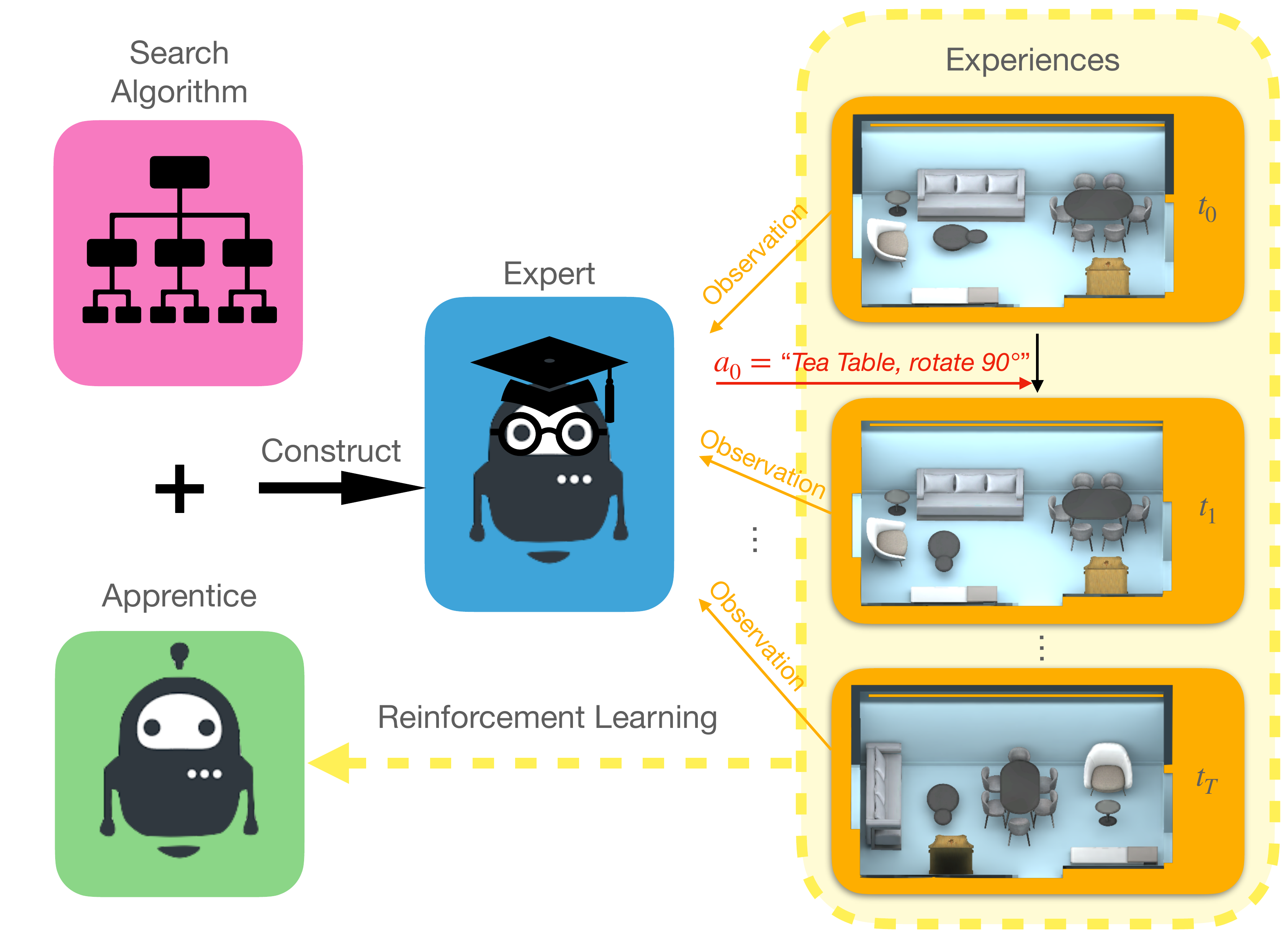}
  \caption{The overview of our framework.}
  \label{fig:overview}
\end{figure}

Although some efforts have been spent on the study of accomplishing
SRP tasks, some critical problems are not solved yet. First, the
definition of the action space is highly constrained. The actions are
designed either to meet specific moving mechanism or to be abstract
and high-level to shorten the length of the decision sequence. With
such coarse-grained or highly constrained action definitions, there
are many unreachable states, which may result in no solution for some
cases. Second, the quantitative evaluation of the agents is limited on
a few real scenes. The evaluation protocol is not convincing enough to
be followed by future work since it covers a limited amount of
situations.

To address these problems, in this paper, we propose a fine-grained
action space definition so that an agent can come up with furniture
moving plans with an arbitrary path in scenes. Thus, an agent can more
likely find an optimal and flexible solution to transform an initial
scene to its target state. More concretely, we define 6 atomic actions
which include four directions of translation and two directions of
rotation to enable flexible and tractable manipulations of furniture
objects.

Training an agent under the atomic action space is challenging because
the use of atomic actions results in longer decision sequences in
general. To tackle this problem, we propose a novel
\textit{Parallelized Expert-Assisted Reinforcement Learning} (PEARL)
framework to efficiently train an agent through self-playing,
leveraging the power of dual policy iteration and reinforcement
learning. In addition, to enable the training of data-hungry learning
approaches and quantitative evaluation, we construct a large-scale
dataset based on 3D-FRONT~\cite{fu20203dfront} which contains $19,062$
pairs of diverse furnished rooms.


Figure~\ref{fig:overview} shows an overview of the proposed PEARL
framework. Under this framework, a search-based expert is constructed
to sample more valuable \textit{state-action} pairs, which alleviate the problem of sparse rewards in the training process. The expert can evolve based on the apprentice policy in each iteration. During the training of apprentice, we take a strategy of reinforcement learning. That is, the apprentice is reinforced by the rewards from the environment and learns from the expert selectively rather than imitating the expert actions directly. The PEARL framework accelerates the learning of the agent and makes the agent converge to a better policy. 

The contributions of our work can be summarized as follows:
\begin{itemize}
\item We propose a more general SRP setting with a flexible and tractable
  action space definition. We also propose a novel PEARL framework to
  efficiently train an agent through self-playing without manual
  annotation and prior knowledge.
\item We introduce a large-scale indoor scene dataset for SRP, which
  contains a diverse variety of realistic, funished room layouts to
  support the training of data-driven approaches and quantitative
  evaluation.
\item We demonstrate the superior performance of the proposed approach
  and analyze the effectiveness of each designed component on the
  introduced SRP dataset.
\end{itemize}

\section{Related Work}
\noindent\textbf{Indoor Scene Synthesis.}
The indoor scene synthesis problem has attracted research attention
for years. It inspired reseach on automatic interior design and
planning serial tasks. With an given empty room, the main goal of
scene synthesis is to coherently predict the pose and position of the
furniture objects to form a realistic layout. Some early works are based
on statistical relationships~\cite{2011Interactive} and
rules~\cite{yu2011make}. The scene layouts are generated through
optimization. Other early works focus on data-driven approaches
for scene synthesis, whereas those approaches are limited by the scale
of training data and available learning methods at that time.

Recent data-driven approaches leverage the power of Deep Learning to
learn from the bigger scene dataset such as
SUNCG~\cite{song2016ssc}. Some works model the spatial relations
between furniture shapes by graphical
models~\cite{Henderson2017Automatic,fu20203dfront}. Some works use
generative approaches~\cite{2018Deep,ritchie2019fast} to model the
distribution of interior layouts and generate new layouts by
sampling. Some other works adopt sequential approach to place
furniture objects iteratively~\cite{2018GRAINS}. With the proposed
generative methods, researchers can easily construct large scale scene
dataset through automatic generation and manual verification, \eg,
3D-FRONT dataset~\cite{fu20203dfront}. The advances of indoor scene
synthesis methods and datasets support further exploration in scene
rearrangement planning.

\vspace{3pt}
\noindent\textbf{Rearrangement Planning.}
Given a certain initial and target configuration of a pre-specified
object set, the Rearrangement Planning problem refers to generating a
feasible action sequence to transform the initial configuration to the
target configuration. This topic has been studied by robotics
researchers while most efforts focus on solving a tabletop rearrangement
task by a robotic arm. The manipulation of the objects can be divided
into two classes: prehensile actions like grasping, picking up and
putting down~\cite{labbe2020monte}, and nonprehensile actions like
pushing.

Many prior works adopt nonprehensile actions since this kind of
manipulations are easier to execute by a robotic arm in practice. In
this direction, the solution of rearrangement planning is to find a
sequence of collision-free move path to push the object to its target
position. Some robotics
works~\cite{haustein2019learning,song2019object,king2016rearrangement,koval2015robust}
solve multi-object rearrangement planning in two stages: a local move
planning stage and a global strategy searching stage, whereas our
approach adopts a universal policy to directly control the move of
objects without isolated planning stages.

Recent graphics work~\cite{wang2020scenem} contributes to the Scene
Rearrangement Planning (SRP) problem. Compared to the tabletop
setting in robotics, scene rearrangement planning is of substantially
higher complexity. For instance, the shapes of objects in a tabletop
setting are usually regular (\eg, cubes, cylinders). Obstacles and
boundaries are usually absent or sparse under a tabletop setting. In
contrast, in realistic indoor scenes, the furniture pieces vary a lot
in shape and dimensions in general; obstacles and boundaries (\eg,
walls, pillars) are typically present. Despite such complexities,
\cite{wang2020scenem} solve SRP by selecting pre-defined paths to move
the objects. Such a coarse-grained action definition is inflexible and
intractable to perform in realistic, complex situations. Different
from the prior SRP setting on the action space, our work adopts a
nonprehensile action space in the problem setting to make the solution
more tractable.

\vspace{3pt}
\noindent\textbf{Dual Policy Iteration.}
Dual Policy Iteration (DPI)~\cite{sun2019dual} algorithms like Expert
Iteration (ExIt)~\cite{anthony2017thinking} and
AlphaZero~\cite{silver2017mastering} have shown impressive performance
on solving decision making problems. This new class of algorithms
maintains two polices: a fast learnable policy (\eg, a neural network)
performs quick rollouts, and a slow policy (\eg, a Tree Search
algorithm) searches the valuable states and plans several steps
ahead. Those two polices are combined to provide superior
demonstrations for the learning of the fast policy, while the updated
fast policy enhances the performance of the combination in return.

In Expert Iteration, the demonstrations from the combined strong
policy are used to supervise the training of the learnable
policy. This supervision is achieved in the form of Imitation Learning
(IL). Different from Expert Iteration, our approach trains the policy
through Reinforcement Learning, which improves the generalizability of
the learnable policy and reduces the learning bias caused by copying
actions of a sub-optimal expert policy.

\section{Preliminaries}
\subsection{Markov Decision Process}
Sequential decision making procedure is often considered in a Markov
Decision Process (MDP). A discounted infinite-horizon MDP is defined
as a tuple
$(\mathcal{S},\mathcal{A},\mathcal{P},\mathcal{R},\gamma)$~\cite{1994Markov},
where $\mathcal{S}$ is the space of \textit{states}, $\mathcal{A}$ is
the space of \textit{actions}; $\mathcal{P}$ is the \textit{transition
  function}: $\mathcal{P}(s'|s,a)$ is the probability of transforming
$s$ to $s'$ by taking action $a$, $s'$ is also written as
$\mathcal{P}(s,a)$; $\mathcal{R}$ is the \textit{reward function}:
$\mathcal{R}(s,a)$ represents the reward received from the environment
by taking action $a$ at the state $s$; and $\gamma$ is the discount
factor. A distribution over the valid actions $a$ given the state $s$
is called a \textit{policy}, denoted as $\pi(a|s)$. The value function
$V^\pi(s)$ is the expectation of accumulated discounted reward by
following $\pi$ starting in state $s$, \ie,
\begin{equation}
  V^\pi(s) = \mathbb{E}_{|s}^{\pi}[\sum_{t=0}\gamma^{t}\mathcal{R}(s_t,a_t)\pi(a_t|s_t)],\ s_0=s.
\end{equation}
The optimal policy $\pi^*$ ought to maximize this
expectation. Formally, we have
$\pi^*=\mathop{\arg\max}_{\pi}V^\pi(s)$.

\subsection{Monte Carlo Tree Search}
Monte Carlo Tree Search (MCTS)~\cite{kocsis2006bandit,guo2014deep} is
a useful strategy to address the challenge of selecting/learning
policies for MDP. It is used to estimate the value of states through
repeated Monte Carlo sampling and simulations. Each node in the search
tree corresponds to a state $s$. The root node represents the current
state. The edge from $s_1$ to $s_2$ represents the execution of action
$a$ by which $s_1$ can be transformed to $s_2$.

In MCTS, four phases are repeated to grow the search tree: a)
selection; b) expansion; c) rollout; and d) back-propagation. In the
selection phase, a feasible unexpanded edge in the tree is selected
according to a \textit{tree policy}. The commonly used tree policy is
the Upper Confidence bounds for Trees (UCT)~\cite{auer2002finite},
\ie,
\begin{equation}
  \text{UCT}(s,a) = \frac{r(s,a)}{n(s,a)} + C\sqrt{\frac{\log n(s)}{n(s,a)}},
\label{equ:uct}
\end{equation}
where $r(s,a)$ is a prior function that gives suggestion on search. A
canonical choice of this function returns a probability of taking
action $a$ at state $t$. It could be estimated by simulations or a
learnable function~\cite{silver2017mastering}, \eg, neural
network. $n(s,a)$ is the counter function which returns the visiting
times of this edge; $n(s)$ is the counter function for the node
corresponding to state $s$; $C$ is the tradeoff parameter to tune the
influence of the terms. In the expansion phase, a new node corresponds
to state $s'\in \{x|\mathcal{P}(x|s,a)\}$, which is expanded from the
selected edge. Then, in the rollout phase, a quick simulation under
the \textit{rollout policy} is performed from the state $s'$ until a
terminal condition is satisfied. The results of simulation are used as
the estimate of the node. Finally, in the back propagation phase, the
information of nodes, including the visited times $n(s)$, $n(s,a)$ and
the estimated value $V(s)$, is updated bottom-up through the tree until
it reaches the root node.

When the tree is built, $a^*=\mathop{\arg \max}_a E(s_{\text{R}},a)$
is the action selected by MCTS, where $s_{\text{R}}$ is the state of
the root node and $E(s,a) = \mathcal{R}(s,a) + \gamma V(\mathcal{P}(s,a))$.

\subsection{Expert Iteration}
The success of employing Deep Reinforcement Learning (DRL) in playing
Atari games~\cite{mnih2015human} brings a great impact to the
field. Various Deep Reinforcement Learning paradigms were proposed to
solve MDP and achieved remarkable performance. However, those methods
suffer from sparse rewards when facing challenging tasks, commonly
resulting in slow convergence and local minima. Those drawbacks can be
alleviated by introducing Imitation Learning (IL). In IL, an agent is
taught to mimic the actions from experts. Experts can arise from
observing humans completing a task and a labeled dataset can be
constructed. Nevertheless, acquiring human demonstrations is
expensive. The agent trained in static data also has difficulty in
generalizing to novel observations.

Expert Iteration provides a method to create an expert which is able
to evolve. The expert is composed of a boosting framework and a fast
vanilla policy (\eg, neural network). The canonical choice of the
boosting framework is a tree search algorithm (\eg, MCTS) which
explores the possible states and exploits the gathered information to
generate a strong move sequence. The vanilla policy can bias the
searching towards more promising states and provide a quick estimation
for the explored state. The vanilla policy is also known as the
\textit{apprentice policy} because it can imitate the actions taken by
the built expert to accelerate the training. The expert can be
improved by embedding the updated apprentice policy.

\section{Method}
\subsection{Problem Definition}
The Scene Rearrangement Planning (SRP) can be modeled as a MDP. The
goal of SRP is to transform the current layout of the scene to the
target layout. An agent can achieve this goal by learning a policy to
make decisions and maximize the accumulated discounted reward during
moving.

\noindent\textbf{State Representation.}
The state in SRP is the scene layout, which includes the shapes,
positions and orientations of the movable objects, and the impassable
districts, \eg, walls. Following~\cite{wang2020scenem}, we use the
top-down view to characterize the scene layout since most movable
objects lay on the floor plane. The objects and impassable districts
are projected on the floor plane, the projection is bounded in a
square and discretized into a $N\times N$ grid
representation. Suppose a scene has $K$ objects, the state of the
scene layout is defined as a binary tensor
$l, l\in \{0,1\}^{N\times N\times (K+1)}$.
   The first two
dimensions of the matrix are the spatial dimensions and the third
dimension is the object dimension. For the $k^{th}$ object, a cell
$l_{i,j,k}$ stores whether the location $(i,j)$ is occupied by the
object $k$. The additional channel of the third dimension is for denoting the
occupancy of the impassable districts. The state $s_t$ at time step
$t$ is the concatenation of the scene layout $l_t$ and the target
scene layout $l_T$ along the third dimension\footnote{$l_t$ and $l_T$
  share the same impassable district channel.},
$s_t \in \{0,1\}^{N\times N\times (2K+1)}$.

\vspace{3pt}
\noindent\textbf{Action Space.}
The action for SRP is defined as a
pair $(o, p), o\in O, p\in P$, where $O$ is the set of objects and $P$
is the set of possible actions which the object can be manipulated
by. Previous work~\cite{wang2020scenem} defined $P$ as a set of
constrained paths to move along, \ie moving straightly along the up,
down, left or right direction until the object reaches an obstacle
(\eg, another object or a wall), and moving along a feasible path
searched from the object's current position to its target
position. Such definition shortens the length of the decision sequence
to limit the complexity of the solution space, making their approach
more focused on moving order planning. This definition has two strong
assumptions: 1) the simulation during path searching is accurate; and
2) the execution of path following is accurate. However, both of the
assumptions are not promised in practice because the information
gathered by sensors and execution of actions are susceptible to
noises. Performing the sophisticated generated actions may result in a
failure.

In robotics, more atomic action spaces are preferred since they are
more tractable. Some robotic works~\cite{yuan2019end,song2019object}
adopt nonprehensile actions (\eg, push) in the rearrangement
problem. Such action definition can be more easily achieved through a
real mechanical arm. In this spirit, we define 6 simple nonprehensile
actions to transform the positions and orientations of the
objects. The first 4 actions are moving an object towards
\textbf{up}, \textbf{down}, \textbf{left} or \textbf{right} direction
by 1 unit. The last two actions are rotating an object clockwise or
anti-clockwise by 15 degrees.

\vspace{3pt}
\noindent\textbf{Reward Shaping.}
When an action is executed, the environment returns a reward value
immediately. Our reward term definitions are shown in
Table~\ref{tab:rewards}. There are 4 kinds of reward term in our
reward shaping. \textit{Distance} returns the change of distance after
an action. Note that the distance includes the Manhattan distance
between the current and target coordinates and the discretized
distance between the orientation states. \textit{Arrival} returns $4$
if the current object arrives at its target state after an
action. \textit{Leave} returns $-4$ if the current object leaves its
target state after an action. \textit{Success} returns a big positive
value $50$ if all objects reached their target states. At each time
step, after an action is executed, the triggered reward terms are
summed up as the immediate reward value.

\begin{table}[t]
  \centering
  \begin{tabular}{llll}
  \toprule
   & \textbf{Reward} &  & \textbf{Reward} \\
  \midrule
  Distance & $\pm 1$ & Arrival & 4      \\
  Success   & 50 & Leave & -4 \\
  \bottomrule
  \end{tabular}
  \caption{Rewards used in SRP.}
  \label{tab:rewards}
\end{table}

\subsection{PEARL}
To solve the challenging problem, we propose a novel variation of the
expert iteration framework to train the agent efficiently. The
boosting framework is a MCTS algorithm. Our agent plays the role of
apprentice, which is composed of a policy network $\pi_\theta$ to
predict the action distribution as well as a value network $v_\phi$ to
predict the future value. $\theta$ and $\phi$ are parameters of the
neural networks.

The expert iteration has two stages in each iteration: 1) Improving
stage and 2) Learning stage. In the improving stage, the apprentice is
used to build the expert and the expert creates a playing record for
the current episode. To construct the expert, the networks are used in
the selection and simulation phases of MCTS. Specifically, in the selection
phase, we use the policy network and the value network to suggest a
more promising action to explore. The formulation of action score can
be derived from Eq.~\ref{equ:uct},
\begin{equation}
  \text{UCT}_\text{NN}(s,a) = \frac{\pi_\theta(a|s)}{n(s,a)} + C\sqrt{\frac{\log n(s)}{n(s,a)}},
\end{equation} 
In the simulation phase, the value of an expanded tree node is
evaluated by the value network. In the learning stage, the policy
network and the value network are trained in a reinforced fashion. In contrast to simply imitating expert actions in expert iteration, the
learning of the apprentice in our method is reinforced by the rewards
from the environment, which means that the apprentice is encouraged or
discouraged to take the expert action accordingly. The learning is similar to Advantage Actor-Critic (A2C)~\cite{mnih2016asynchronous}, where the policy network plays
the role of actor and the value network is the critic. Since MCTS selects
the action by the value of the action that the agent can take at the
root node, it provides both an expert action and an expert estimate of
the state value. The expert estimate of a state value is
$V(s;\theta,\phi)=\max_a \mathcal{R}(s,a) + \gamma
V(\mathcal{P}(s,a);\theta,\phi)$. Then the loss of the policy network
is:
\begin{equation}
  \mathcal{L}_p = -\sum_{t=0} A(s_t,a_t)\log(\pi_\theta(a_t|s_t)),
\label{equ:lp}
\end{equation}
where $A(s_t,a_t)$ is the advantage defined as:
\begin{align*}
  A(s_t,a_t)= &\sum_{i=0}^k \gamma^i\mathcal{R}(s_{t+i},a_{t+i}) + \gamma^{k}V(s_{t+k};\theta,\phi)\\
   &- v_\phi(s_t).
\end{align*}
The value network tries to mimic the state value estimated by the expert. The loss of the value network is:
\begin{equation}
  \mathcal{L}_v = \sum_{t=0} \parallel v_\phi(s_t)-V(s_t;\theta,\phi) \parallel_2.
\label{equ:lv}
\end{equation}

In this way, our
method can potentiallly reduce the bias of learning target on the optimal policy
compared to imitation learning for expert iteration.

\subsection{Training}
We train the networks in a parallelized online synchronous manner,
where a group of data processes synchronize with the latest network
parameters and collect the experiences to support the training of the
network in the optimization process. Since the fast convergence is a well known advantage 
of imitation learning, similar to \cite{wang2019reinforced}, we tradeoff the imitation signal and reinforcement signal to speed up the training. 

It is worth noting that the improving stage is extremely time
consuming compared to the learning stage because of the large
simulations in the improving stage. Actually, in the learning stage,
it is a waste if the apprentice learns from the selected action and
observation records for only once. To improve data utilization, we
adopt Prioritized Experience Replay (PER)~\cite{Schaul2016Prioritized}
in network training.

\vspace{3pt}
\noindent\textbf{Prioritized Experience Replay.}
Experience Replay~\cite{1992Self} is wildly used in online
reinforcement learning. To satisfy the i.i.d. assumption for
stochastic gradient-based algorithm, the collected experiences are
stored in a replay buffer and randomly sampled to feed the
network. Conventional experience replay samples the experiences
uniformly while PER samples the experiences with a priority to take
a significance of the experience into account. PER has been
demonstrated to achieve great performance improvement on many tasks
(\eg, Atari games). In our network training, we prioritize the
experiences with the critic error because the critic error suggests the difference between the value estimated and the value in reality. The high critic error is often caused by novel situations.
 It reinforces the agent to learn from those valuable experiences. We also utilize PER in all baseline agents training.



\begin{figure}[t]
  \includegraphics[width=\linewidth]{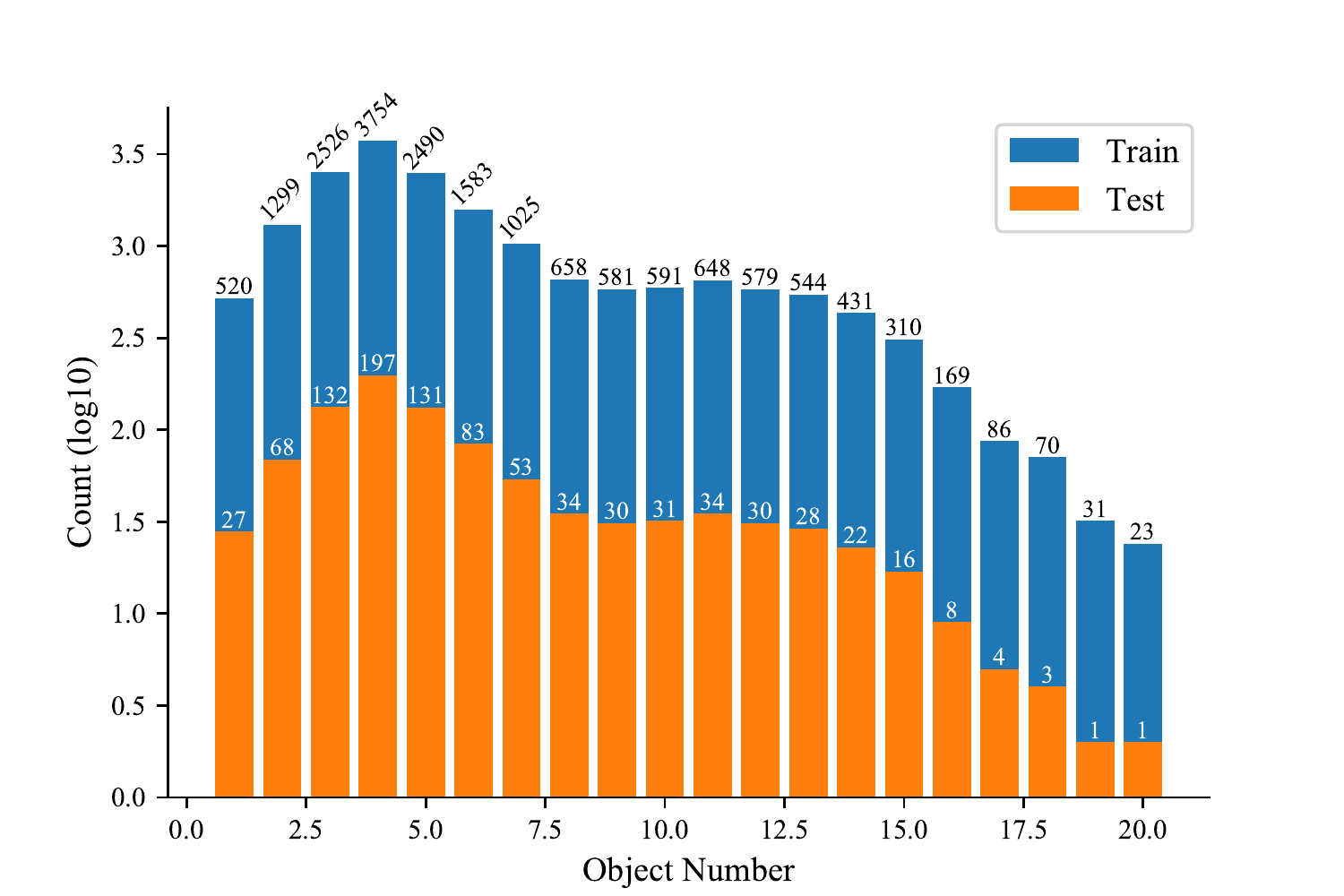}
  \caption{The statitics of the object number in rooms.}
  \label{fig:data}
\end{figure}

\section{Implementation}
\noindent\textbf{Network Architecture.} 
We modify the network architecture from~\cite{wang2020scenem} to build
the actor, which is a CNN encoder appended with a fully connected (FC)
layer. The feature size of the FC layer is $512$. The critic shares
the same CNN encoder with the actor and has an individual FC
layer. The input of the two networks is the current state $s$. The
output of the actor is a $150$-\textit{d} tensor which represents the
predicted probability of actions. It works for at most $20$ objects in
a scene. The output of the critic is a scalar, which refers to the
estimated value of the current state.

\vspace{3pt}
\noindent\textbf{Reproducibility.} 
Our network is implemented in PyTorch. The batch size is $200$.
The number of the data processes is $8$.
The rounds of tree search for each decision making is $50$. In network
training, the maximum allowed action steps in each episode is
$100$. In testing, an episode is regarded as failing if it is not
finished within $200$ steps. The size of the replay buffer is
$10,000$. The optimizer used in the network training is ADAM. The
learning rate is set to $10^{-4}$. The utilities including the virtual
environment and the tree search algorithm are implemented in C++. The
full model is trained on 4 NVIDIA RTX 3090 GPUs with 24GB memory in
each card. For reproducibility, our implementation will be released.

\section{Experiments}
\subsection{Data}

\begin{figure}[t]
  \includegraphics[width=0.9\linewidth]{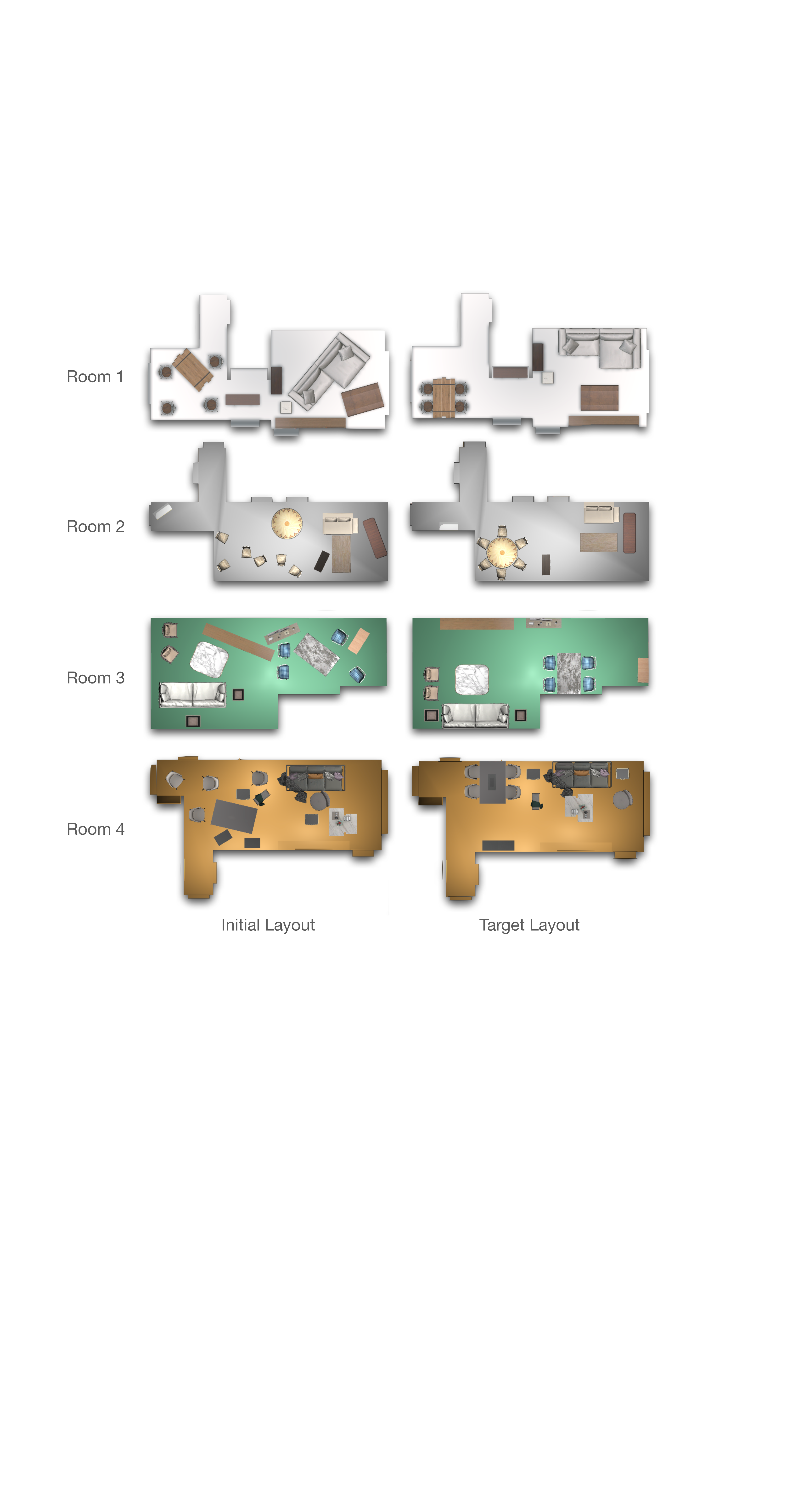}
  \caption{Four cases of scene rearrangement data in our dataset. The initial layout is synthesized through random walks starting from the target layout.}
  \label{fig:room}
\end{figure}
We demonstrate our approach on
3D-FRONT~\cite{fu20203dfront}, a large-scale indoor
scene dataset, where the rooms layout are professionally designed and
populated with high-quality 3D models. The interior designs of the
rooms are transferred from expert creations. This dataset contains
6,813 distinct houses and 19,062 furnished rooms. It is proposed to
support indoor scene tasks like 3D scene understanding, indoor scene
synthesis, semantic segmentations, \etc. The diverse variety of
furnished rooms and the abundant labels satisfy the demand of the SRP
task. In SRP, the complexity of planning highly depends on the number
of objects in the scene. The rooms in 3D-FRONT have 6.1 objects in
average. More than $99\%$ rooms in the dataset have a number of
objects ranging from $1$ to $20$. The statistics of the number of
objects in each room is shown in Fig.~\ref{fig:data}. We split the
dataset into the training and test sets. 5\% of the data (953 rooms)
are randomly sampled as the test set and the rest (18,109 rooms)
belong to the training set.

We use the designed layout as the target layout and randomly sample a
feasible layout with the same objects as the initial
layout. To ensure that the layout pair can be transformed from each other,
 we sample an initial layout by rounds of random
walks starting from the target layout. In each round, a random object
is selected to perform a random feasible move. Each initial layout is
generated by performing 1,000 rounds of random walks. The
configurations of the layouts are extracted from the top-down
sihouette of the scene. We first bound the view of the scene to the
center of a square. Then the objects are rendered individually and the
top-down sihouettes are discretized to extract the shape and position
for each object. The same process is also adopted to extract the
information of impassable districts. The resolution of the discretization is $64\times64$.

\subsection{Quantitative Experiment}
\noindent\textbf{Metrics.}
To evaluate the performance of the agent, we define 2 metrics, \ie,
\textit{Success Rate} $\uparrow$, and
\textit{Length} $\downarrow$, for SRP. Success Rate (SR) is the rate
of attaining the target layout. It illustrates the agent's general
ability of finishing the task. It is the primary metric in our experiment.
 Length refers to the average length of
action sequence. It reflects the efficiency of the action sequence. For the failed
cases, the length of action sequence is the maximum allowed action step (200 in testing).

\vspace{3pt}
\noindent\textbf{Baseline Agents.}
We compare the agent trained by our approach with the following baseline agents:
\begin{itemize}
  \item \textit{RL}: An agent trained with standard A2C. $\epsilon$-greedy strategy is adopted. The actor is used as the agent in inference.
  \item \textit{Expert Iteration}: An agent trained with Expert Iteration. The apprentice actor is used as the agent in inference.
\end{itemize}

For a fair comparison, the networks used in our approach, RL, and Expert Iteration share the same architecture. PER strategy is unitilized in the training of those agents.

\vspace{3pt}
\begin{table}[t]
  \centering
  \resizebox{0.8\linewidth}{!}{
  \begin{tabular}{c|cc}
  \toprule
   & \textbf{SR}$\uparrow$ & Length$\downarrow$ \\
  \toprule
  \cline{1-3}
  RL    & 0.096  &  192.1   \\
  Expert Iteration (Apprentice)    & 0.142  & 177.1     \\
  PEARL (Apprentice) (\textbf{Ours})   & \textbf{0.159} & \textbf{175.4}\\
  \bottomrule
  Expert Iteration (Expert)    & 0.725  &   109.8\\
  PEARL (Expert) (\textbf{Ours})  & \textbf{0.735}  &  \textbf{109.0}\\
  \bottomrule
  \bottomrule
  \end{tabular}
  }
  \caption{The performance of different agents. \textbf{SR} is the primary metric. The number in \textbf{bold} is the best in comparison.}
  \label{tab:quan}
\end{table}
\noindent\textbf{Comparison.}
The comparison between the agents on test sets are shown in
Table~\ref{tab:quan}. The checkpoints of the learnable agents are
selected by SR on test sets. The first three lines 
are the performance of single neural networks.
The apprentice agent trained by Expert Iteration beats the one trained
by RL, which shows the effectiveness of expert actions in
training. The apprentice agent trained by our approach outperforming all baseline agents in all metrics. We also evaluate the expert equipped
with different apprentices. As shown in the last two lines of Table.\ref{tab:quan}, the expert of our approach achieves a high SR of 73.5\% and consistently outperforms the expert of Expert Iteration.

\begin{figure}[t]
  \includegraphics[width=\linewidth]{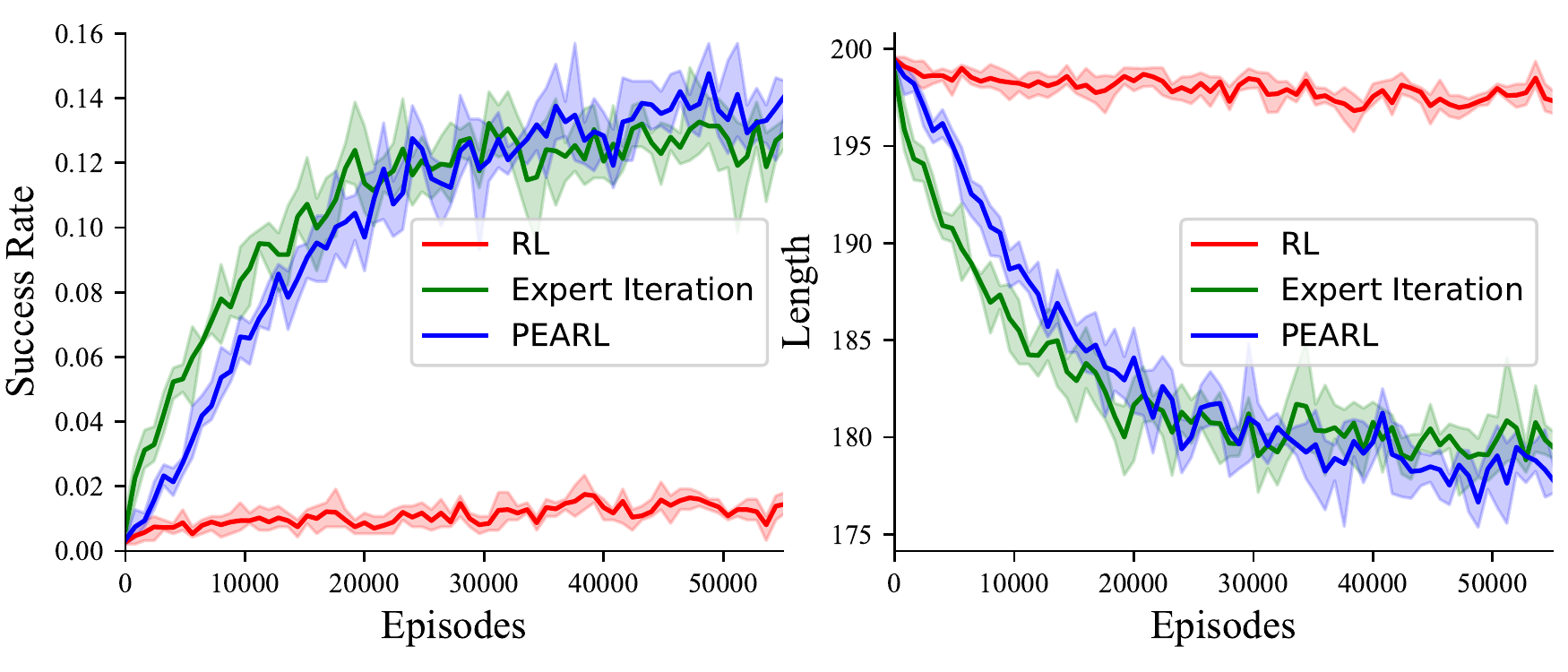}
  \caption{The training curve of the agents.}
  \label{fig:curve}
\end{figure}

To analyze the training behavior of the agents, we plot the curve of SR and Length during training. As shown in Figure~\ref{fig:curve}, the x-axis is the number of epsiodes the agent experienced, the y-axes are SR and Length respectively. Compared to RL, the class of expert-assisted algorithms (\ie, Expert Iteration, and PEARL) learn very fast, which illustrates the high efficiency of the actions sampled by expert. Though the performance of our approach grows slower than Expert Iteration at the beginning, it finally achieves higher success rate than Expert Iteration. It shows that the
actor trained in a reinforced manner converges to a better policy
compared to directly imitating expert actions.




\section{Conclusion}
In this paper, we propose a more general SRP task setting with a
flexible atomic action space definition. To tackle the challenging
problem setting, we propose a novel training paradigm, PEARL, which
leverages the power of dual policy iteration and reinforcement
learning. A large-scale SRP dataset is introduced for training and
evaluating SRP agents. Based on the dataset, we conduct
experiments to validate the proposed approach, The results show that
the agent trained by the proposed learning approach achieves
outstanding performance compared to the baseline agents.

Our approach automates the generation of a move plan for scene
rearrangement and facilitates the realization of scene designs,
leading to potential applications such as warehouse automation, home
rearrangement robots, and smart homes with movable furniture.

Currently, in our problem setting and introduced dataset, we only
consider the rearrangement task on a 2D plane. The furniture objects
can only move and rotate on the floor plane. We do not consider
scenarios with multiple floors nor scenarios with more terrain
constraints. Enriching the SRP task dataset and increasing the variety
of scenes could empower the agent to deal with more complicated
scenarios.

We devise a move planning approach to automate scene
rearrangement. Future works may integrate object-level, robot action
space considerations for plan execution with high-level move
planning. Incorporating factors related to robot movement constraints
into the rewards is a promising extension.

\appendix
\small
\bibliographystyle{../named}
\bibliography{../ijcai21}

\end{document}